\documentclass[preprint,12pt]{elsarticle}

\usepackage{amssymb}
\usepackage{amsmath}

\usepackage{xcolor}

\journal{}

\begin{document}

\begin{frontmatter}



\title{Real-time News Story Identification}


\author[fri]{Tadej Škvorc} 
\author[ijs,mps]{Nikola Ivačič}
\author[kliping]{Sebastjan Hribar}
\author[fri]{Marko Robnik-Šikonja}

 \affiliation[fri]{organization={University of Ljubljana, Faculty of Computer and Information Science}, 
 country={Slovenia},
 email={\\ \{tadej.skvorc,marko.robnik\}@fri.uni-lj.si}
}

 \affiliation[ijs]{organization={Jožef Stefan Institute, Department of Knowledge Technologies}, 
 country={Slovenia},
 email={\ nikola.ivacic@ijs.si}
}
\affiliation[mps]{organization={Jožef Stefan International Postgraduate School}, 
 country={Slovenia},
}
 \affiliation[kliping]{organization={Kliping d.o.o.}, 
 country={Slovenia},
  email={\ sebastjan.hribar@kliping.si}
}


\begin{abstract}
To improve the reading experience, many news sites organize news into topical collections, called stories. In this work, we present an approach for implementing real-time story identification for a news monitoring system that automatically collects news articles as they appear online and processes them in various ways. Story identification aims to assign each news article to a specific story that the article is covering. The process is similar to text clustering and topic modeling, but requires that articles be grouped based on particular events, places, and people, rather than general text similarity (as in clustering) or general (predefined) topics (as in topic modeling). We present an approach to story identification that is capable of functioning in real time, assigning articles to stories as they are published online. In the proposed approach, we combine text representation techniques, clustering algorithms, and online topic modeling methods. We combine various text representation methods to extract specific events and named entities necessary for story identification, showing that a mixture of online topic-modeling approaches such as BERTopic, DBStream, and TextClust can be adapted for story discovery. We evaluate our approach on a news dataset from Slovene media covering a period of 1 month. We show that our real-time approach produces sensible results as judged by human evaluators.
\end{abstract}




\end{frontmatter}



\section{Introduction}
\label{sec:introduction}

The ever-increasing amount of global news presents an overwhelming challenge for individuals and organizations striving to keep abreast of relevant information. 
The sheer volume of unstructured text data makes it nearly impossible for users to consume and comprehend all the information pertinent to their interests. 
News monitoring and analysis systems can help us by employing advanced algorithms to address the problems of size and opinionated views, as well as misinformation. 
Sensible grouping of news, such as identifying common reported events or clustering, allows users to quickly glean essential information from a set of articles, saving time and reducing cognitive load.

\begin{figure}[h]
\centering
\hspace*{-1.3in}
\includegraphics[width=0.6\textwidth]{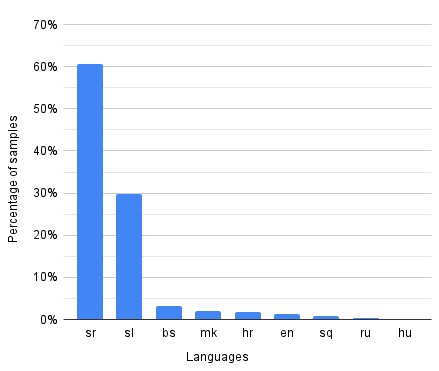}
\caption{The distribution of languages in all samples (Serbian, Slovene, Bosnian, Macedonian, Croatian, English, Albanian, Russian, and Hungarian).}
\label{fig:samples_lang_hist}
\end{figure}

The overall targeted news monitoring system archive comprises over 85 million articles from eight countries (Serbia, Slovenia, Bosnia and Herzegovina, Croatia, Macedonia, Montenegro, Kosovo, and Albania) in more than seven languages (see Figure~\ref{fig:samples_lang_hist}) spanning over twenty years. 

The current system performs dynamic clustering over short time intervals, leveraging text embeddings generated by an information retrieval Transformer encoder. 
This process operates on result sets defined by client-selected queries and filters.
While this approach offers advantages, such as the ability to cluster a specific subset, adjust clustering thresholds per query, and choose between different embedding models, it also presents notable limitations. 
Specifically, the resulting clusters are unstable over time and are constrained by the size of the retrieved result set, resulting in inconsistent and arbitrary clustering outcomes.

Our objective is to develop a method for global, archive-wide clustering that remains independent of individual client searches (i.e., static concerning query context), while continuously expanding over time as new articles are added to the archive. 

To address this problem, our approach is anchored in clustering of articles based on reported events, a process we refer to as story identification. 
We model this as the task of identifying news stories that persist over time and grouping related articles accordingly. 
This approach enables us to construct a coherent, high-level view of the extensive archive, supporting meaningful analysis of the development and impact of specific events.

The main contribution of our work is the creation of a story identification system that combines different text representation methods, named entity recognition, summarization, hierarchical clustering, and topic modeling to identify news stories in real time, as they appear in the targeted media monitoring system. Additionally, we present a comprehensive evaluation showing how different components of the story identification system impact the results.

The structure of our paper is as follows. In Section \ref{sec:related_work}, we present an overview of past work related to clustering, topic modeling, and text representation methods. Section \ref{sec:dataset} contains the description of the dataset used for our research, as well as an overview of several problems that make this task more challenging than general text clustering or classification. We follow with an overview of the developed methodology in Section \ref{sec:methodology}. We first present our approach to optimizing text representation for story identification, followed by an overview of our approach to online clustering. Section \ref{sec:results} contains the evaluation of our approach, comparing different components of our algorithm. We conclude our work in Section \ref{sec:conclusion} and present directions for further work.

\section{Related work}
\label{sec:related_work}
We split the overview of the related work into four subsections, presented below. We present related work on clustering, topic modeling, and general text representations. We end the section with a discussion on the problems of existing approaches for story identification.

\subsection{Clustering and online clustering}
Although not an established task in the field of natural language processing, story identification can be viewed as a subset of text clustering. Text clustering aims to generate clusters that contain similar texts, while story identification aims to generate clusters of articles addressing the same news story. Many general-purpose approaches for clustering have been proposed in the past, generally aiming to minimize differences between texts that occur within the same cluster and maximize distances of texts from different clusters. Most often, this is tackled through iterative minimization algorithms such as K-means \cite{macqueen1967some}, which minimizes inter-cluster variance, or density-based algorithms such as DBSCAN \cite{ester1996density} and MEAN SHIFT \cite{cheng1995mean}. More complex approaches can also be used, such as modeling clusters using Gaussian mixtures \cite{yang2012robust}, which allow for more complex cluster representations (i.e., modeling them as Gaussian distributions rather than simple cluster centers).

While such algorithms may be employed for general story identification, our use case requires identifying stories in real-time, as the articles are published. General clustering approaches require the entire dataset to be present in advance, making them incompatible with this task. Instead, several approaches capable of online clustering have been proposed. For example, STREAMKmeans \cite{o2002streaming} modifies the original k-means algorithm to perform clustering iteratively, DBSTREAM \cite{hahsler2016clustering} performs online clustering using a shared density graph in combination with a variant of the DBSCAN algorithm, and CluStream \cite{aggarwal2003framework} first groups points into micro clusters that are merged into larger clusters using k-means after a set number of points have been processed. However, such approaches are generally not fine-tuned for story identification and may not take into account important information (e.g., the fact that articles from the same story need to refer specifically to the same events, people, and places).

\subsection{Story identification and topic modeling}
Story identification is a more specific task than general-purpose clustering. Instead of clustering texts with similar content, we want to identify texts that describe the same news story or event. This means that two texts with relatively similar content can describe a completely different event. For example, a news article describing a sports event taking place in Paris would probably have similar content to one describing a sports event in Berlin, even though the event is different. Likewise, articles describing a presidential election may describe the same event from various points of view, resulting in articles with limited text similarity. Therefore, clustering approaches that work on general text similarity may be unsuitable for our task. To address this, we either need more specific clustering approaches or text representation methods capable of emphasizing the important parts of each news story.

More specific clusters can be obtained using topic modeling. Topic modeling builds upon clustering algorithms with additional steps aimed at identifying topics present in a given text. This can range from simple dimensionality reduction algorithms aimed at detecting keywords in each cluster (e.g., LDA \cite{blei2003latent}) to more comprehensive methodologies. For example, \citet{grootendorst2022bertopic} presents BERTopic, which combines document embeddings, dimensionality reduction, clustering, and additional fine-tuning into a more complex topic modeling approach. Complex approaches can be better at capturing keywords that are important for story identification (e.g., places, people, and events), but it is difficult to say whether their topics correlate specifically with story identification.

\subsection{Text representation approaches}
Both clustering and topic modeling approaches rely on fundamental text representation methods (i.e., how we transform text into vectors that are necessary for algorithmic analysis). Many different representation models have been proposed. In their simplest form, the bag-of-words approach \cite{harris1954distributional} constructs text based on word counts. More complex approaches use machine learning to train embedding models that are more capable of taking into account the context of words \cite{mikolov2013efficient}. Current state-of-the-art methods build upon this idea by using large language models based on the contextual transformer architecture \cite{vaswani2017attention} to produce text embeddings that efficiently capture the content of a given text \cite{devlin2019bert}.

\subsection{Problems with existing approaches}
General text representation models may prove insufficient for story identification. News stories place a high importance on specific people, places, and events and less importance on the general meaning of a given text. Unless a model has been trained to detect these specifics, it is unlikely to prioritize them sufficiently. Additionally, many text-representation models are designed with short texts in mind (e.g., sentences) and perform poorly when applied to longer pieces of text. For example, approaches such as Sentence-BERT \cite{reimers-2019-sentence-bert} perform best on sentences or small paragraphs, not entire articles.

In recent years, large language models such as GPT-4 \cite{achiam2023gpt}, DeepSeek \cite{liu2024deepseek}, and LLAMA \cite{touvron2023llama} have become increasingly popular, both for text representation and other tasks related to natural language processing, several of which may be of use for story identification. For example, text summarization can be employed to reduce the length of an article (removing issues with text representation for longer articles), while information retrieval can be used to extract specific places, people, or events.
 
To address the issues of existing approaches, we propose a new method that uses fine-tuned online topic modeling in combination with custom text representation methods. We base our approach on online BERTopic and DBStream algorithms, fine-tuning the clustering parameters and introducing fixes and constraints specifically aimed at story identification. We combine this with a text representation approach that was designed to prioritize components that are relevant to story identification while placing less weight on the overall text content. We also explore whether summarization using large language models can be integrated into the pipeline to improve results. We provide an in-depth explanation of our method in Section \ref{sec:methodology}.

\section{Data}
\label{sec:dataset}

\begin{figure}[htb]
\centering
\hspace*{-1.3in}
\includegraphics[width=0.6\textwidth]{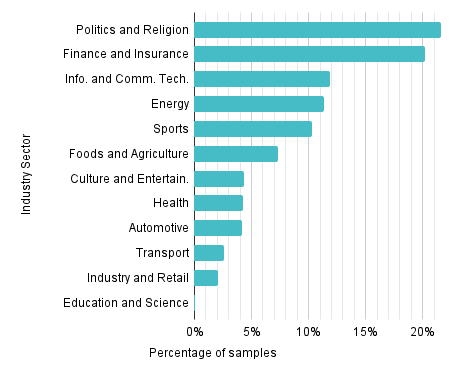}
\caption{The distribution of selected industry sector labels across our sample.}
\label{fig:sectors_samples}
\end{figure}

To develop and evaluate our method, we utilized data provided by a leading media monitoring company that maintains an extensive archive of news articles.
To ensure a tractable problem scope, we sampled a subset of articles from a large-scale news monitoring archive.
The selection process was guided by industry-sector labels, enabling comprehensive coverage of sector-specific news (see Figure~\ref{fig:sectors_samples}) while also maintaining representative linguistic diversity of the region’s media landscape.

To manually evaluate the quality of story detection produced by the current system's dynamic clustering method, we further down-sampled the dataset. 
Specifically, we focused on Slovene articles published between March 01 and April 01, 2023, and defined six discrete time frames for evaluation: March 01, 06, 11, 16, 21, and 26. 
This language and temporal segmentation allowed us to assess clustering behavior and story formation within constrained windows. 
Each news article data sample consisted of the following fields:
\begin{itemize}
    \item The country and name of the media outlet.
    \item The publication and processing dates as recorded by the media monitoring company.
    \item The title of the news article.
    \item The full body text of the article.
    \item Manually assigned industry-sector labels provided by the media monitoring company (e.g., \textit{banking}).
    \item Text embeddings precomputed using OpenAI's \textbf{text-embedding-ada-002} model.
\end{itemize}

To identify semantically coherent clusters of articles for manual evaluation, we replicated the current system's dynamic clustering method: a multi-stage pipeline consisting of embedding extraction, pairwise similarity computation, graph-based community detection, and post-hoc cluster aggregation.
The method is summarized as follows:
\begin{enumerate}
    \item \textbf{Embedding Extraction}:
    
    Let $\mathcal{A} = \{a_1, a_2, \ldots, a_n\}$ be a collection of articles, and $n$ their cardinality. For each article $a_i \in \mathcal{A}$, we produce a dense semantic embedding vector $e_i \in \mathbb{R}^d$ combining the news article title and body text representation.
    
    \item \textbf{Similarity Computation}:
    
    We compute the pairwise cosine similarity between all article embeddings, resulting in a similarity matrix $S \in \mathbb{R}^{n \times n}$, where $S_{ij} \in [-1,1]$. A binary adjacency matrix $A \in \{0,1\}^{n \times n}$ is constructed by thresholding $S$ using a predefined threshold $\tau \in \mathbb{R}$, such that
    \[
    A_{ij} =
      \begin{cases}
        1 & \text{if } S_{ij} > \tau \\
        0 & \text{otherwise}
      \end{cases}
    \]

    \item \textbf{Graph Construction and Community Detection}:

    A graph $G = (V, E)$ is constructed from the adjacency matrix $A$, where each node $v_i \in V$ corresponds to an article and an edge $(v_i, v_j) \in E$ exists if $A_{ij} = 1$. Louvain modularity-based community detection~\cite{Blondeletal2008} is applied to identify clusters of densely connected nodes, optimizing community structure for a resolution parameter $\gamma = 0.1$.

    \item \textbf{Cluster Aggregation and Ranking}:

    We reconstruct clusters by grouping articles with identical community labels. 
    The resulting clusters are sorted in descending order based on cluster cardinality. 
    Within each cluster, articles are ranked by media outlet reach (in descending order) to identify a representative article. 
    The top-ranked article in each cluster is used as a news story prototype.
\end{enumerate}
This process was iteratively refined by adjusting key hyperparameters, such as the similarity threshold and resolution parameter, and manually evaluated by the media monitoring company until the resulting clusters achieved a satisfactory structure. Finally, the produced dataset was manually re-evaluated, restructured, and re-ordered to represent the final evaluation dataset.

To efficiently evaluate our method, we focused on articles between March 03 and March 13, giving us a dataset of 6400 articles covering 4028 different news stories. This provides us with a time window of 10 days, corresponding to the time window of our clustering method, described in Section \ref{sec:methodology}. We use this dataset for automatic analysis of the proposed methods as it contains the ground-truth data necessary for such tests.

Several aspects of our data make it challenging to cluster with existing approaches. In the labeled dataset, the average length of each article is 3525 characters or 550 words. This means that many articles surpass the maximum token limit of many standard embedding approaches (e.g., BERT, which has a limit of 512 subword tokens). Methods that support longer inputs may struggle to focus on the relevant parts of articles.

\begin{figure}[htb]
\centering

\includegraphics[width=\textwidth]{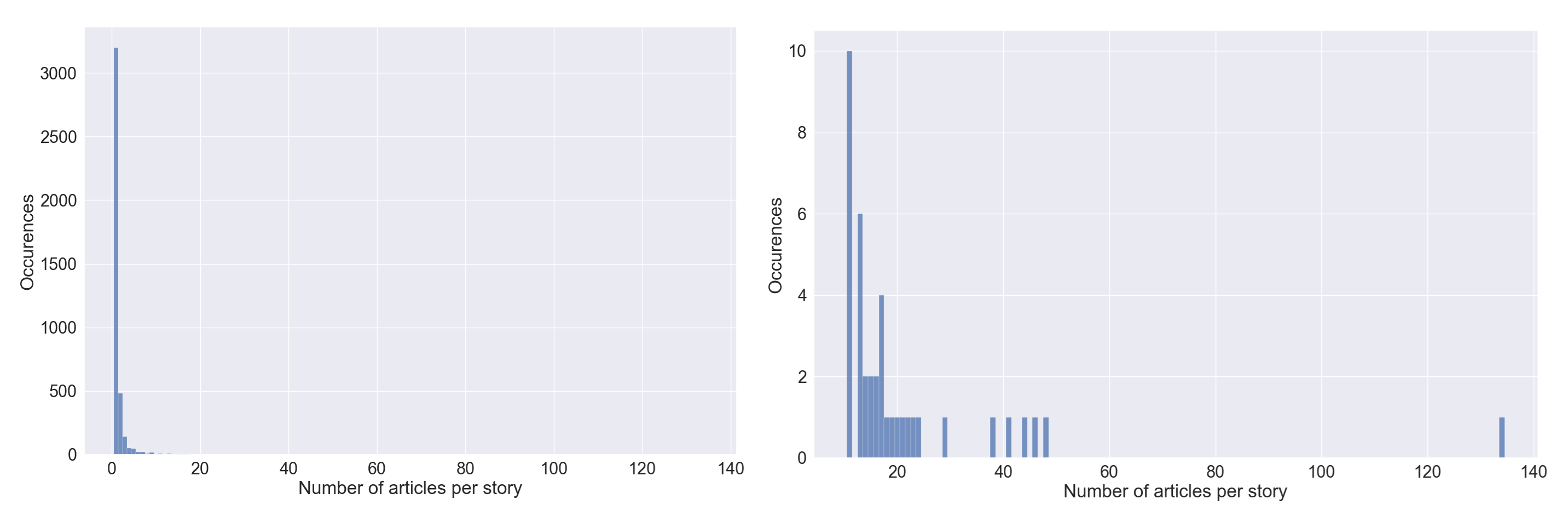}
\caption{The distribution of story sizes in our labeled dataset. The left side of the graph shows the distribution for all stories. The right side of the graph shows the distribution for stories with 10 articles or more.
}
\label{fig:dist_story_sizes_all}
\end{figure}

Another issue is the distribution of story sizes. Our labeled dataset contains 6400 articles split into 4028 stories. A significant number of stories, therefore, contain only a small number of articles, often as few as one article per story. Figure \ref{fig:dist_story_sizes_all} shows the distribution of the story sizes. Most stories (3199) contain only a single article. This is followed by a large number of stories (789) that contain between 2 and 10 articles. Only 40 stories contain more than 10 articles, the largest containing 134 articles. 

This skewed distribution is a natural result of both news publishing (important stories tend to dominate the news cycle with a large amount of articles, while smaller stories may only get a single article) and the small time period of our labeled dataset (a span of 9 days means stories near the beginning or end of the time period will contain a smaller amount of articles). Thus, the described phenomenon places greater importance on two aspects of clustering:
\begin{itemize}
    \item \textbf{Detecting outliers.} Since stories with a small number of articles (e.g., 1 or 2) represent a majority of our data, care must be taken to properly identify them and avoid grouping them with similar clusters.
    \item \textbf{Handling both large and small stories.} The differences in story sizes make it difficult to set appropriate limits for cluster sizes. To correctly assign news articles to stories, the approach must strike a good balance between small and large cluster sizes.
\end{itemize}

We propose an approach that handles the above specifics and describe it in Section \ref{sec:methodology}.

\section{Story Identification Methodology}
\label{sec:methodology}
\begin{figure}[htb]
\centering
\includegraphics[width=0.8\textwidth]{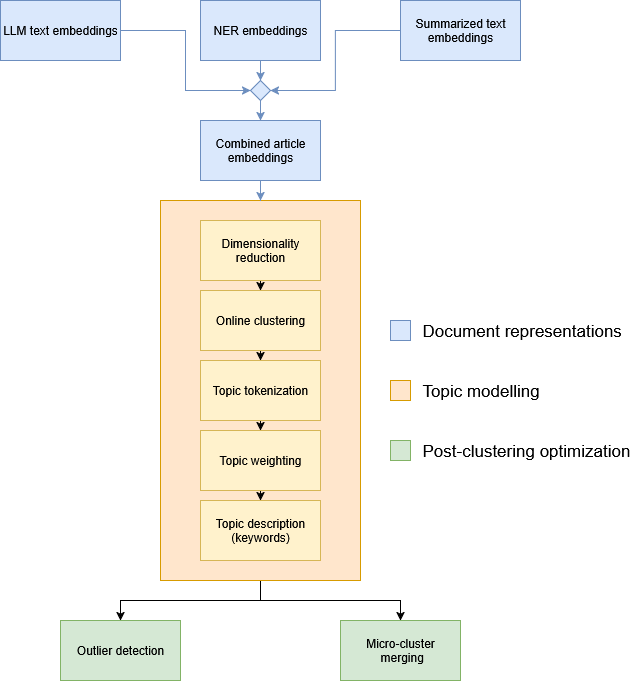}
\caption{The schematic overview of the proposed approach to news story identification.}
\label{fig:algo_diagram}
\end{figure}

An overview of our methodology is presented in Figure \ref{fig:algo_diagram}.
In Section \ref{subsec:text_representation}, we present our approach to text representation. We focus on capturing specific people, places, and events that are the most important factors in story identification and describe how we tackle the representation of long texts. In Section \ref{subsec:online_clustering}, we describe how we cluster news articles in real time, i.e., making our approach fast enough to handle real-world data. In Section \ref{subsec:post_clustering_optimization}, we describe additional optimization that can be performed after online clustering to improve the final results. 

\subsection{Text Representation for Story Identification}
\label{subsec:text_representation}
To perform text clustering and story identification, we must first convert each article from text to vector form. As discussed in Section \ref{sec:dataset}, traditional text embeddings are inadequate for our task, as they miss specific aspects necessary for story identification (i.e., mentions of particular events, people, or places) or might be unable to deal with lengthy news articles. In order to address these issues, we experiment with different text embedding methods.

\subsubsection{Text embeddings using LLM-based approaches}
In order to deal with longer text lengths, we compare several state-of-the-art LLM-based text embedding approaches, including DistilBERT \cite{Sanh2019DistilBERTAD} and TSDAE \cite{wang2021tsdae} with models that are capable of working with longer text lengths (e.g., BGE-m3 \cite{bge-m3}, with a context window of 8192 tokens). These generally provide good representations of the articles' content. However, they may have issues identifying important aspects of story identification (e.g., placing enough emphasis on mentions of specific people or events). To address this problem, we combine them with additional forms of embeddings.

\subsubsection{Explicit named entity recognition}
While LLM-based embeddings capture the general content of each news article, they may not be separate stories that have similar textual content but describe entirely different events. For example, a story describing a football match between team A and team B might have a very similar text to one describing a match between team C and team D. A general text clustering would likely put these articles into the same cluster. However, story identification needs to identify them as two different stories.

To ensure this, we rely on named entity recognition (NER) using the sloner model \cite{prelevikj2023pytorch}. NER models are explicitly trained to detect named entities (e.g., persons, locations, organizations, and places). Named entities alone are not enough for story identification, but can be helpful when combined with other text representations.

\subsubsection{Vectorizing Summarized Articles}
Another way to extract important information from articles is text summarization. In recent years, LLM-based text summarizers have become increasingly valuable, with multiple models specialized in news summarization. These models should be able to summarize an article in a few sentences while keeping the most important aspects of each article intact. This eliminates the issue with article length and implicitly performs named entity recognition: since key named entities are important aspects of a news article, a good summarizer should keep them in a summary.

Ideally, an automatic summarizer would summarize the entire news article into a short paragraph while preserving the information necessary for story identification. We test a summarizer specifically trained on Slovene news articles \cite{vzagar2023one} to determine whether this approach can help with story identification.

\subsubsection{Combining Text Representations}
While each text representation method embeds the article into a vector, each method alone is unlikely to capture every aspect of the article (i.e., LLM-based embeddings may miss certain named entities, while NER embeddings disregard everything that is not a named entity). To overcome this, individual representations can be combined by concatenating their vectors. The final representation vector is obtained as a concatenation of vectors produced by the approaches described above.

We experiment with a combination of embedding approaches to determine which combinations perform best. The results are presented in Section \ref{sec:results}.

\subsection{Clustering}
\label{subsec:online_clustering}
While a text representation is necessary for story identification, it is not sufficient. A key challenge is that stories need to be identified in real time (i.e., as they appear in the media monitoring system). An additional request for clustering is to take into account the publishing time of news articles.

Standard clustering approaches operate on the entire dataset at once. Due to time constraints, running the whole clustering algorithm each time a new article appears in the system is not feasible. To address this, we experiment with different online clustering approaches and introduce additional constraints based on articles' publishing times.

\subsubsection{Online clustering}
We base our approach on existing online clustering algorithms, specifically DBSTREAM (a general-purpose online clustering algorithm) and TextClust \cite{assenmacher2022textual} (an online clustering algorithm specialized for text clustering). Both approaches operate by first grouping points into micro-clusters that are later merged into final macro-clusters. This is important for story identification because it allows us to control the size of each cluster (story) by controlling the merges into macro clusters. We describe this in more detail below in Section \ref{subsec:post_clustering_optimization}.

Additionally, both mentioned algorithms can take into account the time a story was published when assigning it to a cluster. They implement a fading factor $\lambda$ that reduces the importance of past articles. We describe how this can be used to improve story identification in Section \ref{subsec:time-based constraints}.

Online clustering can also be implemented in topic modeling approaches. We use a well-known topic detection approach, BERTopic \citet{grootendorst2022bertopic}, and replace its clustering algorithms with DBSTREAM and TextClust. We then apply time-based constraints and post-clustering optimization immediately after the clustering.

\subsubsection{Time-based Constraints}
\label{subsec:time-based constraints}
A significant difference between text clustering and story identification is the presence of time-based information (i.e., when a given story was published). Using the historical data and information provided by Kliping, we assume that a story contains articles from at most 10 days. This practical limitation allows stories to capture real-world events that span multiple days but prevents a single story from stretching over a longer time period.

We implement this time constraint by setting the fading factor $\lambda$ so that it approaches zero if a cluster contains 10-day-old articles, ensuring that articles outside the time period have weights that are low enough to not influence the clustering steps.

\subsubsection{Hierarchical Topic Modeling and Post-clustering Optimization}
\label{subsec:post_clustering_optimization}
To further improve the clustering results, we perform additional post-clustering optimization, relying on the micro-clusters produced by TextClust and DBSTREAM online clustering algorithms. Usually, the micro-clusters are automatically joined into larger macro-clusters using an approach similar to hierarchical clustering. However, this approach has several disadvantages that we address through the following optimization steps:

\textbf{Outlier Detection}. As outlined in Section \ref{sec:dataset}, a significant amount of our dataset consists of stories with only a single article. Such stories include regional or specific stories only reported in a single news article. For example, the dataset contains several stories related to e-sports that were only covered by a single article on a single news site (Esport1.si). While TextClust and DBSTREAM are capable of detecting outliers, the large number of outliers can be problematic. To avoid it, we first manually detect outliers using a distance-based threshold (i.e., marking an article as an outlier if it is sufficiently distant from all other articles) based on the text and named entities of each article. As shown in Section \ref{sec:results}, using named entities helps differentiate between different stories occurring in similar contexts (e.g., two different sports matches).

\textbf{Additional micro-cluster merging}. Automatic macro-cluster assignment sometimes fails to merge micro-clusters that belong to the same story. This can happen if two stories describe the same event using sufficiently different wording, resulting in a too large difference in text similarity. We manually cluster similar micro-clusters if they contain similar named entities and were published in the same 10-day time window.

We also experiment with cluster merging and separation based on keywords generated by topic modeling. The approach is very similar to our named entity approach, except that the named entities are replaced by the topic modeling keywords.

Due to real-time speed requirements, we do not perform these additional steps after every clustering step. Instead, we do batches of these steps after a day's worth of articles. This ensures that our approach can run in real time.

\textbf{Combining clustering approaches}.
Our final story identification algorithm is the combination of all the steps described above. Figure \ref{fig:algo_diagram} shows a general overview of our algorithm, starting with text representation, continuing with online clustering and topic modeling, and ending with the post-clustering optimization. We experiment with different combinations of steps (e.g., various text representations, clustering methods, and post-clustering optimizations). The results of these methods are presented in Section \ref{sec:results}.

\section{Results}
\label{sec:results}
We evaluate our approach on two datasets described in Section \ref{sec:dataset}. First, we compare text embedding and clustering approaches using the smaller labeled dataset. This allows us to automate the evaluation, using the story labels as ground-truth data. Additionally, we evaluate our approach using unsupervised measures that are commonly used in clustering evaluation (e.g., the Silhouette score, which measures the cohesion of the generated clusters).


We start our evaluation using several commonly used offline clustering algorithms. Due to time constraints described in Section \ref{sec:methodology}, these are not suitable for use in a real-time system, but can provide a simple and uniform way to evaluate individual components of our approach (e.g., the text representation methods).

We test the following approaches and some of their combinations.
\begin{itemize}
    \item \textbf{Text embeddings:} tfidf \cite{sparck1972statistical} (baseline) , DistilBERT \cite{Sanh2019DistilBERTAD}, BGE-M3 \cite{bge-m3}, distiluse-base-multilingual-cased-v2 \cite{reimers-2019-sentence-bert}, paraphrase-MiniLM-L6-v2 \cite{reimers-2019-sentence-bert}, multilingual-e5-large-instruct \cite{wang2024multilingual}, newsqa-msmarco-distilbert-gp (DistilBERT finetuned on the MS MARCO dataset \cite{bajaj2016ms}.
    \item \textbf{Named entities:} no named entities, named entity recognition using sloner \cite{prelevikj2023pytorch}.
    \item \textbf{Online clustering:} DBSTREAM \cite{hahsler2016clustering} , TextClust \cite{assenmacher2022textual}.
    \item \textbf{Topic modeling:} BERTopic \cite{grootendorst2022bertopic} with a KeyBERT \cite{grootendorst2020keybert} representation model.
\end{itemize}

\subsection{Text representation}
Due to the recent proliferation of large-language models, there exist several embedding models suitable for our task. We evaluated several LLM-based approaches and used TF-IDF weighted document vectors as a baseline. We evaluated our approach on a small human-labeled dataset. Due to the large number of outliers (stories that contain only a single article) in our dataset, we evaluated outliers and the rest of the dataset separately. For outliers, we first performed outlier detection using a distance threshold-based approach, where we marked an article as an outlier if it was a sufficient distance away from its nearest article. The distance threshold was calculated using a development set, consisting of a random 10\% sample from the small labeled dataset. On this dataset, we calculated the average outlier and non-outlier distance to the nearest article and set the threshold as the mean of the two distances. For the test set, we performed a balanced sampling between outliers and non-outliers so that the test set contained a 50/50 split of outliers and non-outliers. 

We evaluated the non-outliers using k-means clustering. This test is not necessarily indicative of the final performance, but it allows for a quick comparison of different embedding approaches. We computed the adjusted mutual information (AMI) score as well as the Silhouette score using ground-truth labels. The results are presented in Table \ref{tab:results_embedding}.

\begin{table}[htb]
\centering
 \begin{tabular}{c c c c} 
 \hline
 Embedding method & AMI & Silhouette & Outlier CA  \\  
 \hline
 TF-IDF & 0.603 & 0.204   & 0.724\\ 
 distiluse & 0.777 &  0.376&  0.693\\
 MiniLM-L6 & 0.626 & 0.264& 0.623\\
 TSDAE & 0.662 & 0.239 & 0.544 \\
 DistilBERT & 0.724 & 0.316 & 0.675\\  
 BGE-M3 &  \textbf{0.8378 }&  \textbf{0.445} & 0.720\\
 \hline
 \end{tabular}
 \caption{Comparison of embedding models using k-means clustering and comparing assigned clusters with human labels of stories. Columns present adjusted mutual information (AMI), Silhouette score, and classification accuracy (CA) of outliers.}
 \label{tab:results_embedding}
\end{table}

All LLM-based embeddings outperform the baseline TF-IDF based on the clustering AMI measure, with BGE-M3 obtaining the best results according to both the AMI and the Silhouette score. TF-IDF representation performs well on outlier detection but is not effective at clustering articles. For this reason, we use BGE-M3 as the selected representation method. While the clustering results are not a perfect match to the graph-based method described in Section \ref{sec:dataset}, clustering with BGE-M3 embeddings comes reasonably close and still produces sensible stories. Additionally, the key advantage of the clustering approach is that it can be adapted to function in real time and can therefore be incorporated into a real-time media monitoring system, where the graph-based approach is too time-consuming. Additionally, it does not require manual evaluation and strict hyperparameter tuning.

Next, we investigate the impact of named entity recognition on text embedding performance. We extracted named entities from text using the sloner model and embedded them using the same model as for text, i.e., BGE-M3. We present results of using only the named entities and the named entities concatenated with the text vector in Table \ref{tab:results_ner}. The results show that using NER slightly reduces the general clustering performance but increases the accuracy of outlier detection; on this task, using only named entities outperforms using only the text. 

The results indicate that named entities are significant for outlier detection. Articles covering unique stories are also likely to contain unique named entities (e.g., names of places or people that do not appear in any other article) and can therefore be identified using NER. However, the entire text is necessary to cluster articles containing similar named entities correctly. This means that an optimal clustering approach shall first perform outlier detection using named entities and then cluster non-outliers without using named entities.

\begin{table}[htb]
\centering
 \begin{tabular}{c c c c} 
 \hline
Method & AMI & Silhouette & Outlier CA \\  
 \hline
  BGE-M3 (without NEs) & \textbf{0.8378} &  \textbf{0.445} & 0.720\\
 BGE-M3 (just NEs) &  0.678 &   0.273 & 0.762\\
  BGE-M3(text +  NEs) & 0.791 &  0.375 & \textbf{0.768}\\
 \hline
 \end{tabular}
 \caption{Comparison of text representation with and without named entities (NEs) using k-means clustering and comparing assigned clusters with human labels of stories. }
 \label{tab:results_ner}
\end{table}

Finally, we investigate whether summarization can improve text representation. Ideally, a summarization model would condense the article into a shorter form, maintaining the information necessary for story identification. We present the results in Table \ref{tab:results_summarizer}. The evaluation shows that using summaries instead of full texts decreases the clustering performance across all embedding models (see Table \ref{tab:results_embedding}) but provides a slight improvement in outlier detection (still less than explicitly using named entities). It seems that the summarizer omits critical information for story identification. As it also significantly increases the runtime, we did not use it in further evaluations. 

\begin{table}[htb]
\centering
 \begin{tabular}{c c c c} 
 \hline
Method & AMI & Silhouette & Outlier CA\\  
 \hline
 TF-IDF &  0.650 &  0.242   &  0.709\\ 
 distiluse & 0.693 & 0.220& 0.698\\
 MiniLM-L6 & 0.600 & 0.150& 0.628\\
 TSDAE & 0.570 &  0.121 & 0.577\\
 DistilBERT & 0.610& 0.171 & 0.62\\  
 BGE-M3 &  \textbf{0.780} &  \textbf{0.263} & \textbf{0.730}\\
 \hline
 \end{tabular}
 \caption{Comparison of embedding models on summarized texts using k-means clustering and comparing assigned clusters with human labels of stories.}
 \label{tab:results_summarizer}
\end{table}

\subsection{Online clustering and topic modeling}
After evaluating the embedding methods, we assess the best approach (BGE-M3) together with online clustering approaches. For this analysis, we first select 10\% of the articles as an initial set and cluster them using the chosen online clustering method. We then simulate a real-time news system by iteratively adding batches of 10 news articles (sorted by time) and assigning them to story clusters (or, if necessary, creating a new story cluster) using the chosen online clustering method.

The results of this evaluation are presented in Table \ref{tab:results_online}. All evaluation scenarios use BGE-M3 as the embedding model, as it achieved significantly better results than other models. The outlier detection method remains the same as in Table \ref{tab:results_embedding}. The main benefit of this approach is that it is suitable for real-world use, where re-running the entire clustering algorithm every time a new article is published is infeasible.

\begin{table}[htb]
\centering
 \begin{tabular}{c c } 
 \hline
Method & AMI \\  
 \hline
 DBSTREAM & 0.162\\
 DBSTREAM with BERTopic & 0.398 \\
 TextClust & 0.172  \\  
 TextClust with bertopic & 0.355  \\ 
 \hline
 \end{tabular}
  \caption{Comparison of online clustering approaches by comparing assigned clusters with human labels of stories.}
 \label{tab:results_online}
\end{table}

The results show that online clustering and topic modeling approaches significantly underperform compared to offline clustering. In the following subsection, we check if the results can be improved using post-clustering optimization.

\subsection{Post-clustering optimization}
The online clustering methods we tested, DBSTREAM and TextClust, perform hierarchical clustering, where individual points are first grouped into micro clusters, which are then merged into the final "macro" clusters. However, a key disadvantage of these online clustering methods is that determining the optimal threshold for grouping macro-clusters (i.e., a threshold that will produce the desired number of macro-clusters) can be difficult to determine and needs to be fine-tuned for different scenarios (e.g., news from different languages, or news from different topics may require different thresholds). Since online clustering processes each added point individually, without being aware of future points, they may create macro clusters prematurely, resulting in multiple clusters that should belong to the same story. Conversely, if the clustering threshold is set too high, the online algorithms might avoid merging clusters that belong to the same story.

\begin{figure}[htb]
\centering
\includegraphics[width=1\textwidth]{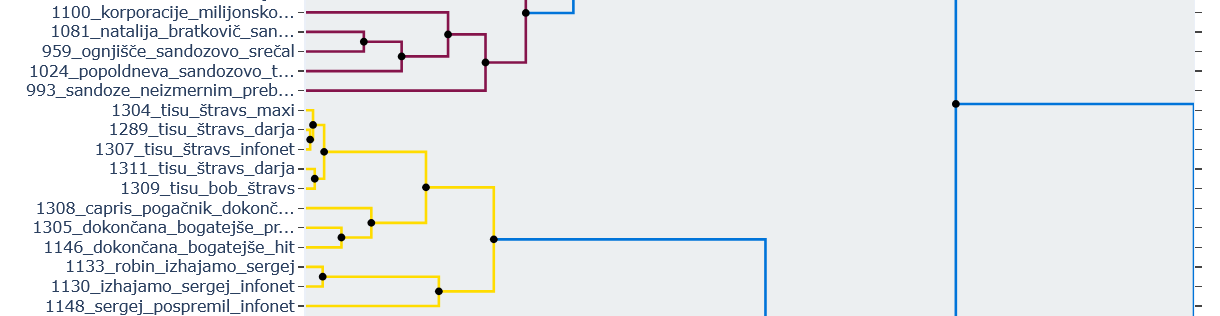}
\caption{An example of macro clusters obtained through iterative clustering and topic modeling. The labels represent topic descriptions generated by the  BERTopic approach. The dendrogram displays the distances between the clusters.}
\label{fig:topic_examples}
\end{figure}

Figure \ref{fig:topic_examples} presents an example where online clustering creates multiple macro clusters belonging to the same story. The Figure shows a dendrogram of cosine distances between cluster centers, where each cluster is described with topic keywords obtained through topic modeling. The clusters displayed in yellow all belong to a story concerning the Slovene photographer Darja Štravs Tisu and contain nearly identical topic keywords, but were not correctly merged into larger macro clusters. This could be addressed by modifying the parameters of online clustering; however, fine-tuning them is challenging because different stories may require different values.

We address this issue through "post-clustering optimization", which involves merging the clusters outside the steps of online clustering. We developed a threshold-based method that repeatedly merges clusters that are closer than a threshold $d$ to each other. Unlike online clustering, where distances are calculated using embedded texts and named entities, we can merge clusters using embeddings of cluster centers or embeddings of key topic terms identified using topic modeling. We generate topic terms from either the entire text or the named entities present in a given text. Table \ref{tab:results_cluster_merging} compares the results of both strategies. 

\begin{table}
\centering
 \begin{tabular}{c c c c} 
 \hline
 & \multicolumn{3}{c}{Merging strategy} \\
Method  & none & topic terms &  cluster centers  \\  
 \hline
 DBSTREAM Text & 0.398 & 0.509 & 0.359 \\
 DBSTREAM NEs & 0.385 & \textbf{0.569} & 0.346\\  
 DBSTREAM Text + NEs& 0.376 & 0.545 &  0.355 \\  
  TextClust Text & 0.355 & 0.555 & 0.415 \\
 TextClust NEs & 0.304 & 0.414 & 0.304\\  
 TextClust Text + NEs& 0.324 & 0.390 &  0.340 \\  
 
 \hline
 \end{tabular}
  \caption{Comparison of online clustering approaches using different post-clustering optimization approaches, i.e., different merging strategies. We report the AMI score. }
 \label{tab:results_cluster_merging}
\end{table}

While both merging scenarios improve the clustering performance, using topic modeling terms outperforms using cluster center embeddings. Additionally, calculating topic terms on named entities improves the results compared to calculating them on the entire text. Named entities are likely more useful for merging clusters because they are better at identifying the key facts of each story compared to topic terms extracted from the whole text. While the performance still lags behind clustering the entire dataset, this approach has a crucial advantage: it is feasible for a real-time application.



%



\section{Conclusion and further work}
\label{sec:conclusion}
In our work, we presented an online clustering system fine-tuned for story identification. We investigated several text embedding methods and suggested improvements to online clustering to show how existing approaches can be improved to better suit story identification. Our evaluations show improvements when using named entity recognition, explicit outlier detection, and optimization to micro-cluster merging.

While we evaluated our approach on a relatively large dataset, further evaluation in a practical setting shall determine how well the developed approach scales in real-world use, particularly in systems that process large amounts of data for longer periods of time. As natural language processing technologies are rapidly advancing, periodic reassessment of new embedding approaches might bring further benefits.

\section*{Acknowledgments}
The work was primarily supported by the Slovene Research and Innovation Agency (ARIS) project L2-50070, and also by the core research programme P6-0411 and project GC-0002. The work was also supported by EU through ERA Chair grant no. 101186647 (AI4DH) and cofinancing for research innovation projects in support of green transition and digitalisation (project PoVeJMo, no. C3.K8.IB). The computational resources were provided by SLING through project S24O01-42.

\bibliography{cas-refs}
\bibliographystyle{plainnat}
\end{document}